\documentclass[10pt,twocolumn,letterpaper]{article}

\usepackage{3dv}
\usepackage{times}
\usepackage{epsfig}
\usepackage{graphicx}
\usepackage{amsmath}
\usepackage{amssymb}
\usepackage{booktabs}
\usepackage{pdfpages}
\usepackage{multido}

\usepackage[pagebackref=true,breaklinks=true,letterpaper=true,colorlinks,bookmarks=false]{hyperref}

\usepackage{pifont}
\newcommand{\cmark}{\ding{51}}%
\newcommand{\xmark}{\ding{55}}%

\threedvfinalcopy 

\def\threedvPaperID{xx} 

\ifthreedvfinal\fi
\usepackage{times}
\usepackage{epsfig}
\usepackage{graphicx}
\usepackage{amsmath,amssymb}
\usepackage{bm,xspace}
\usepackage{verbatim}
\usepackage{multirow}
\usepackage{wrapfig}
\usepackage{balance}
\usepackage{multicol}        
\usepackage{mathrsfs}
\usepackage{mathtools}
\usepackage{bbm}
\usepackage{times}
\usepackage{enumerate}
\usepackage{algorithm} 
\usepackage{tabularx}
\usepackage{array}
\usepackage{bm}
\usepackage{cleveref}
\usepackage[bottom]{footmisc}

\def\eg{\emph{e.g}\onedot} 
\def\ie{\emph{i.e}\onedot}

\def\sota{state-of-the-art}

\newcommand\norm[1]{\left\lVert#1\right\rVert}

\def\xx{\mathbf{x}}

\begin{document}

\title{A Divide et Impera Approach for 3D Shape Reconstruction from Multiple Views}

\author{Riccardo Spezialetti\thanks{Work done while at Google.} $^{1}$, David Joseph Tan$^{2}$, Alessio Tonioni$^{2}$, Keisuke Tateno$^{2}$, Federico Tombari$^{2,3}$\\
		\small $^{1}$ Department of Computer Science and Engineering (DISI), University of Bologna\\
		\small $^{2}$ Google Inc.\\
		\small $^{3}$ Technische Universit\"at M\"unchen\\
}
%
%

\makeatletter
\g@addto@macro\@maketitle{
	\begin{figure}[H]
		\setlength{\linewidth}{\textwidth}
		\setlength{\hsize}{\textwidth}
		\vspace{-5mm}
		\centering
       \includegraphics[width=1.0\linewidth]{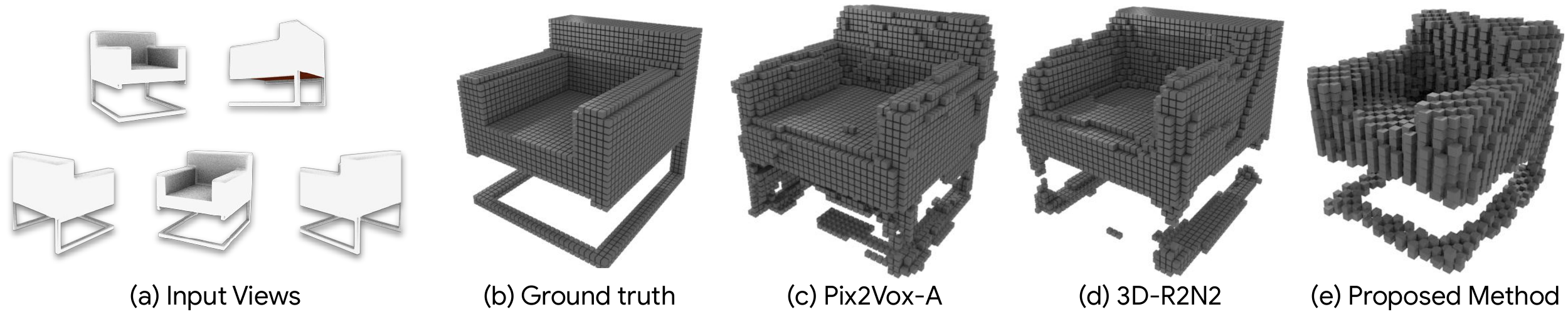} 
		\caption{3D Shape Reconstruction results on the ShapeNet dataset \cite{chang2015shapenet}. Our method successfully reconstructs the base of the chair that is visible in the input views (right part of the image) while Pix2Vox-A \cite{Xie_2019_ICCV} and 3D-R2N2 \cite{choy20163d} fail.}
		\label{fig:teaser}
	\end{figure}
}
\maketitle

\begin{abstract}
Estimating the 3D shape of an object from a single or multiple images has gained popularity thanks to the recent breakthroughs powered by deep learning. 
Most approaches regress the full object shape in a canonical pose, possibly extrapolating the occluded parts based on the learned priors.
However, their viewpoint invariant technique often discards the unique structures visible from the input images.
In contrast, this paper proposes to rely on viewpoint variant reconstructions by merging the visible information from the given views. 
Our approach is divided into three steps.
Starting from the sparse views of the object, we first align them into a common coordinate system by estimating the relative pose between all the pairs. 
Then, inspired by the traditional voxel carving, we generate an occupancy grid of the object taken from the silhouette on the images and their relative poses. 
Finally, we refine the initial reconstruction to build a clean 3D model which preserves the details from each viewpoint.
To validate the proposed method, we perform a comprehensive  evaluation on the ShapeNet reference benchmark in terms of relative pose estimation and 3D shape reconstruction.
\end{abstract}

\section{Introduction}
\label{sec:introduction}

\begin{figure*}[!ht]
	\centering
	\includegraphics[width=0.95\linewidth]{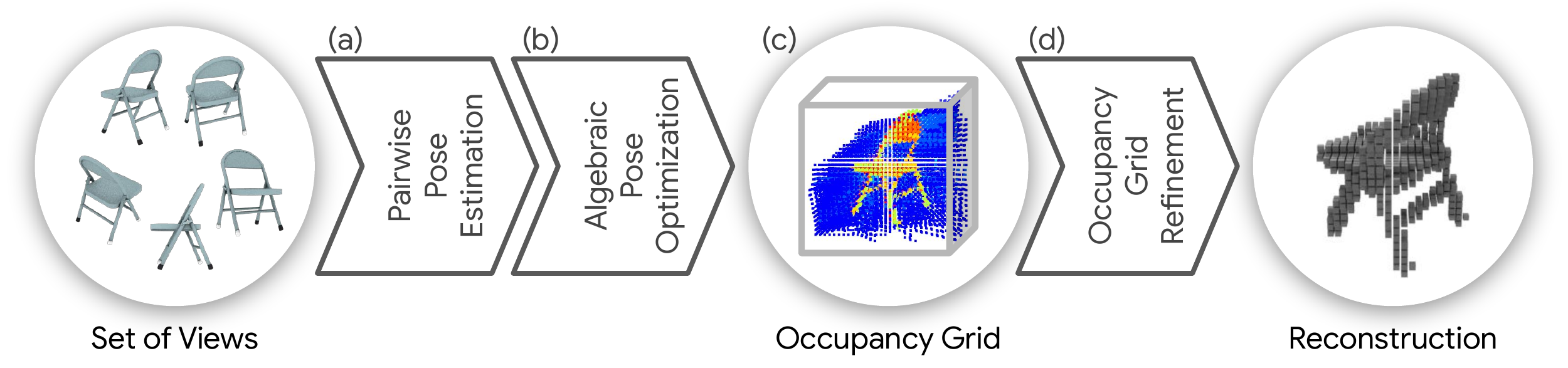}
	\caption{Given a set of views of the object, our framework (a)~estimates the relative poses between pairs of images; (b)~algebraically optimize the pose of each image; (c)~build an occupancy grid from poses and silhouettes; and, (d)~refine the occupancy grid to reconstruct the full model.}
	\label{fig:pipeline}
\end{figure*}

The knowledge of the 3D structure of objects in our surroundings is a necessary information for perception systems aiming to accomplish tasks such as object interaction and modeling. 
A solution to these tasks plays a key role in applications being developed -- or envisioned -- in the field of virtual and augmented reality, robotic manipulation and grasping, rapid prototyping/reverse engineering, and autonomous driving. 
Although humans can guess the 3D shape of an object  from a single glance, this is a complicated task for a computer that requires the ability to estimate the pose of an object and to reconstruct its 3D shape. 
The presence of several, sometimes non-overlapping, images sparsely taken of an object is a common situation in scenarios such as e-commerce, where products are advertised with randomly acquired images and large object databases already exist. 
Existing works can be divided into two main categories -- \textit{single-view 3D object generation} and \textit{multi-view 3D object reconstruction}.
While the latter is a well-known traditional computer vision problem taking into account overlapping views around the same object, the former has only been  recently  addressed successfully thanks to the availability of large 3D CAD model repositories~\cite{chang2015shapenet,lim2013parsing,sun2018pix3d,xiang2014beyond} and to the introduction of deep learning based methods.
Many recent proposals rely on a neural network to reconstruct the entire shape of an object given a single image~\cite{Groueix_2018_CVPR,mescheder2019occupancy,Richter_2018_CVPR,wang2018pixel2mesh,yan2016perspective}.
Although the predicted shapes are visually compelling, different structures can correspond to the occluded part of an object, as examined in \cite{wu2018learning}; hence, determining the full 3D shape from a single view is an ill-posed problem. 
Recent works even question if the task being learned is shape estimation or simply shape retrieval \cite{tatarchenko2019single}.

Multi-view 3D object reconstruction has been addressed in the past by a long line of geometry-based reconstruction techniques such as {structure from motion} (SfM) \cite{ozyecsil2017survey}, {multi-view/photometric stereo} (MVS) \cite{hartley2003multiple} and {simultaneous localization and mapping} (SLAM) \cite{fuentes2015visual}. 
All of these works leverage the insight that seeing the same points of an object or a scene from different perspectives lets us understand its underlying geometry. 
Although these methods are successful in handling many scenarios, they all rely on feature matching that can be difficult when multiple viewpoints are separated by large baselines and the amount of textured regions is low. 
Practically, they require continuous overlap between consecutive views acquired by an educated user (\eg{} moving slowly and avoiding holes).

Recent deep-learning based methods have been proposed to overcome these limitations~\cite{arsalan2017synthesizing,choy20163d,kar2017learning,tatarchenko2016multi,wei2019conditional,wen2019pixel2mesh++,Xie_2019_ICCV}.
Most of them rely on the assumption of either employing the ground-truth camera poses while inferring shapes \cite{arsalan2017synthesizing,choy20163d,kar2017learning,tatarchenko2016multi,wen2019pixel2mesh++} or predicting models oriented in a canonical reference frame \cite{choy20163d,deng2019cvxnets,Xie_2019_ICCV}.
But neither the camera pose nor the canonical reference frame are readily available for an arbitrary object. 
Moreover, both single and multi-view reconstruction methods tend to learn as a prior an average structure of a category of objects from the training set. 
As a result, when the input object deviate significantly from the average structure, \eg{} a chair with less than four legs (\eg, \autoref{fig:teaser}-a), the reconstruction fails badly by trying to make it fit to the average shape.
Differently, we combine deep learning with traditional computer vision wisdom to reconstruct the objects and explicitly take into account the geometric structures that are visible in the image but are not common in the average structure, as illustrated in \autoref{fig:teaser}. 
To do so, we propose a novel shape generation using color images captured from a limited number of views with unknown camera poses and in arbitrary reference systems. 
The complete pipeline is shown in \autoref{fig:pipeline}.
Our method first constructs a set of all possible pairwise views and estimate the relative poses between them through a CNN-based pose estimation.
We then rearrange the pairs in a {fully-connected graph} and convert it to a over-determined system that can be further refined by an algebraic pose optimization.
Subsequently, we rely on the estimated poses to build the first rough approximation of the 3D object reconstruction by projecting the object silhouettes from every views to a shared occupancy grid.
Finally, we refine the initial occupancy map with a 3D CNN to produce the final voxelized model estimation. 

To the best of our knowledge, we are the first to propose a method for multi-view shape generation from non-overlapping views without the need of the poses or without relying on the canonical orientation of the model.
\section{Related work}
\label{sec:related}

\paragraph{Single-view 3D Shape Generation.}
Single-view shape generation methods hallucinate the invisible parts of a shape by implicitly memorizing the statistical distribution of the training data. 
A central role for the quality of the estimated shape is played by the representation chosen to encode it. 
One of the more common choice is to regress the shapes as 3D voxel volumes \cite{choy20163d,girdhar2016learning,hane2017hierarchical,kar2015category,tulsiani2017multi,wu2017marrnet,wu2016learning,yan2016perspective}
while other 3D parametrization are indeed possible such as 
octrees~\cite{riegler2017octnet,tatarchenko2017octree,wang2017cnn}, 
point clouds~\cite{Fan_2017_CVPR,insafutdinov2018unsupervised,lin2018learning,mandikal20183d}, 
mesh~\cite{chen2019bsp,Groueix_2018_CVPR,Kato_2018_CVPR,wang2018pixel2mesh}, 
depth images~\cite{arsalan2017synthesizing,Richter_2018_CVPR,tatarchenko2016multi}, 
classification boundaries~\cite{chen2019learning,mescheder2019occupancy} 
and signed distance function~\cite{park2019deepsdf}. 
Generation approaches were also explored for single-view reconstruction including geometric primitive deformation \cite{Groueix_2018_CVPR,kanazawa2018learning,kurenkov2018deformnet,tulsiani2016learning,wang2018pixel2mesh}, a combination of Generative Adversarial Networks (GANs) \cite{goodfellow2014generative} and Variational Autoencoders (VAEs) \cite{kingma2013auto} in \cite{wu2016learning}, and 
re-projection consistency~\cite{kato2018neural,tulsiani2018multi,tulsiani2017multi,yan2016perspective}. 
However, all the aforementioned works are based on the assumption that the non-visible parts of the object can be hallucinated given the shape priors learned from the training data. 
Unfortunately, this assumption does not always holds since the occluded parts may not have a deterministic shape, as examined in \cite{wei2019conditional,wu2018learning}. 
Consequently, the inferred shapes tend to suffer from over smoothed surfaces without fine details. 
Contrary to these methods, we rely on a viewpoint-variant modelling in order to learn a representation that keeps and fuse together the visible geometric details from each viewpoint. 

\paragraph{Multi-view 3D Shape Generation.}
Extracting the 3D geometry from multiple views is a well-researched topic in computer vision. 
Methods based on SfM \cite{brown2005unsupervised,ozyecsil2017survey,snavely2006photo,ullman1979interpretation},
MVS \cite{hartley2003multiple}
and SLAM \cite{fuentes2015visual} require to match features across images. 
But the matching process becomes difficult when the viewpoints are separated by wide baselines and when the images are poorly textured. 
The dawn of deep learning has fostered several methods for multi-view 3D reconstruction --
multi-view geometric cues are exploited when training single-view prediction architectures as a supervisory signal \cite{insafutdinov2018unsupervised,lin2018learning,rezende2016unsupervised,tulsiani2018multi,yan2016perspective}, or when training and inferring  multi-view systems to enrich the learned representation by capturing view-dependent details \cite{choy20163d,kar2017learning,wei2019conditional,wen2019pixel2mesh++}. 
Unfortunately, recovering this information across-views is hard if the camera poses are unknown since these methods follow the assumption that the camera pose of each image is given.

Kar \etal{} \cite{kar2017learning} learn a model for multi-view stereopsis \cite{kutulakos2000theory}. 
In \cite{wen2019pixel2mesh++}, a coarse shape generated as in \cite{wang2018pixel2mesh} is refined iteratively by moving each vertex to the best location according to the features pooled from multiple views. 
But both methods require the exact ground truth camera poses. 

A different way of avoiding the need for camera poses is to reconstruct the object from each view in a canonical reference system and let the network learn the view-dependent shape priors.
Choy \etal{} \cite{choy20163d} propose a unified framework to create a voxelized 3D reconstruction from a sequence of images (or just a single image) captured from uncalibrated viewpoints. 
As soon as more views are provided to the network, the reconstruction is incrementally refined through a {Recurrent Neural Networks} (RNNs).
However, due to the permutation-variant nature of RNNs, the results can be inconsistent and, in addition, are time-consuming.

An alternative way, explored in \cite{wang20173densinet}, is to pool only maximum values. 
To overcome these limitations in \cite{Xie_2019_ICCV}, they introduce a learned context-aware fusion module that acts on the coarse 3D volumes generated on each input images in parallel.
In \cite{wei2019conditional},  Wei \etal{} propose to learn a generative model conditioned by multiple random input vectors. 
With different random inputs, they can predict multiple plausible shapes from each view to overcome the ambiguity of the non-visible parts. 
The final model is obtained by taking the intersection of the predicted shapes on each single-view image.
Similar to our work, the method proposed in \cite{wang2019deep} first predicts a coarse shape together with the object pose from an image using CNN then refines the coarse volume with the help of a neural network. 
Differently, in our approach, the input of the refiner is constructed from the visual cones defined by different views. Furthermore, our pose estimator predicts the relative pose between the pairs of views instead of regressing the absolute one.

\paragraph{Shape and pose recovering.}
Researches have also proposed methods for simultaneous 3D shape reconstruction and camera pose estimation from a single-view \cite{insafutdinov2018unsupervised,sun2018pix3d,tulsiani2018multi}. 
Tulsiani \etal \cite{tulsiani2018multi} regress multiple poses hypothesis for each sample, forcing the learnt distribution to be similar to a prior distribution to deal with the ambiguity of the unsupervised pose estimation. 
In contrast, Insafutdinov \etal{}~\cite{insafutdinov2018unsupervised} train an ensemble of pose regressors and use the best model as a teacher for knowledge distillation to a single student model.
Finally, in \cite{sun2018pix3d}, the pose estimation is treated as a classification problem by discretizing the azimuth and elevation angles.

\newcommand{\langular}{\mathcal{L}_\text{angular}}
\newcommand{\lcontours}{\mathcal{L}_\text{contours}}
\newcommand{\lpose}{\mathcal{L}_\text{pose}}
\newcommand{\rotation}{R}
\newcommand{\rtrue}{R^*}
\newcommand{\ttrue}{t^*}
\newcommand{\roptimized}{\hat{\rotation}}
\newcommand{\rpredicted}{\tilde{\rotation}}
\newcommand{\quaternion}{q}
\newcommand{\qtrue}{\quaternion^*}
\newcommand{\qtrueinv}{{\quaternion^{*}}^{-1}}
\newcommand{\qpredicted}{\tilde{\quaternion}}
\newcommand{\qoptimized}{\hat{\quaternion}}
\newcommand{\shapepredicted}{p_\Delta}
\newcommand{\shapetrue}{p^*}
\newcommand{\lrefiner}{\mathcal{L}_\text{refiner}}
\newcommand{\voxbintrue}{p_{i}^*}
\newcommand{\voxbinpred}{\tilde{p}_{i}}
\newcommand{\voxcellpred}{\tilde{p}_{i, j, k}}
\newcommand{\voxcelltrue}{p_{i, j, k}^*}
\newcommand{\pcdpredicted}{\tilde{P}}
\newcommand{\pointspcdpredicted}{\{\tilde{\xx}_n\}}
\newcommand{\pcdtrue}{P^*}
\newcommand{\pointspcdtrue}{\{\xx^{*}_n\}}

\newcommand{\image}{I}
\newcommand{\imageset}{\mathcal{I}}
\newcommand{\source}{\image_s}
\newcommand{\target}{\image_t}
\newcommand{\poseNetwork}{\mathscr{N}}
\newcommand{\resnet}{ResNet-18}
\newcommand{\contour}{\mathcal{V}}
\newcommand{\contourPoint}{v}
\newcommand{\distanceTransform}{\mathcal{D}}
\newcommand{\quaternionset}{\mathcal{Q}}
\newcommand{\silhouette}{S}
\newcommand{\silhoutteset}{\mathcal{S}}

\renewcommand{\rotation}{\mathbf{R}}
\newcommand{\translation}{\mathbf{t}}

\begin{figure*}[!t]
\centering
    \includegraphics[width=0.95\linewidth]{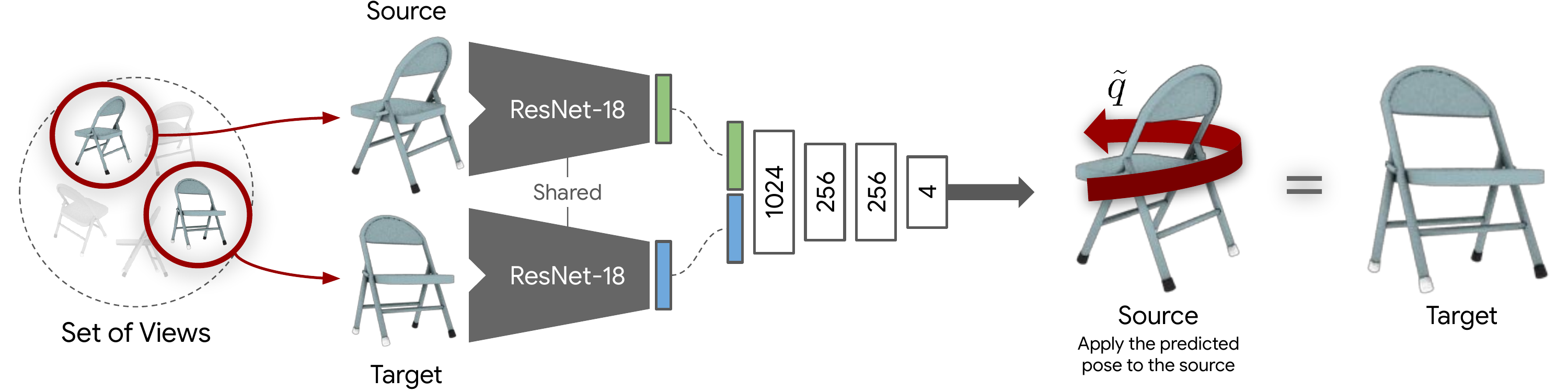}
    \caption{Schematic representation of our relative pose estimation architecture. A \resnet{} based encoder that extracts image descriptors on both the source and target images. These two embeddings are then concatenated and examined by a fully-connected network that outputs the relative rotation between the input images.}
	\label{fig:pose_estimation}
\end{figure*}

\section{Proposed method}
\label{sec:method}

Given a set of $N$ images acquired from different viewpoints around the same object $\imageset = \{\image_i\}_{i=1}^{N}$, the objective is to reconstruct the 3D shape of the object.
The proposed method is three-fold:
(1)~estimation of the relative pose between every permutation of the image pairs in $\imageset$ by means of a task-specific model (see \autoref{subsec:pose_estimation});
(2)~graph-based rectification of the bidirectional relative poses to refine them and calculate the absolute poses with respect to a reference image (see \autoref{subsec:pose_optimization}); and,
(3)~reconstruction based on building an occupancy grid from the image silhouettes and the estimated poses to then refine it through a 3D CNN (see \autoref{subsec:occupancy_grid}).

\subsection{Relative pose estimation}
\label{subsec:pose_estimation}
Recovering the 3D geometry of an object from a set of images requires them to be correlated either through their relative or absolute poses.
The absolute poses have the disadvantage of requiring a canonical reference coordinate for every object.
The canonical system needs to be unique for each object class and is usually handpicked, which imposes a strong constraint to the system that might not be optimal for many use cases. 
On the other hand, predicting the relative poses between all pairs of views removes this constraint.
Therefore, influenced by the {classical} 3D reconstruction frameworks, in our pipeline, we regress the relative poses between a couple of views instead of the absolute ones. 
Unlike classical 3D reconstruction, however, our approach does not require overlapping regions between the two images. 

We start by arranging the set of input images in a fully connected graph by establishing the links of all the possible permutations of image pairs.
We then estimate the relative pose considering one image as the source ($\source$) and every other image as the target ($\target$).
Defining the pose as a unit quaternion $\qpredicted \in \mathbb{R}^4$, we aim at learning a deep neural network function $\poseNetwork{}$ such that $\poseNetwork{}(\source, \target) \rightarrow \mathbb{R}^4$.
A schematic representation of the architecture is depicted in \autoref{fig:pose_estimation}.
We employ a siamese \resnet{} to extract a 1024-dimensional deep embedding from each view independently.
Then, we concatenate the two representations and elaborate the resulting 2048-dimensional vector with a stack of fully connected layers to finally regress the pose at the last layer.  

\subsubsection{Contour loss}
We generate the training set from the 3D model of the object.
Following \cite{insafutdinov2018unsupervised,tulsiani2018multi}, images are created by rendering the model according to the camera projection matrix $\pi$ from random views with a fixed translation $\ttrue$. By dividing them into pairs, ground truth relative pose $\qtrue$ from $\source{}$ to $\target{}$ are produced.
We adapt the contour loss from \cite{manhardt2018deep} to supervise the regression of the pose from the network.
To find the contours, we project the point cloud to the target image and sample a sparse set of points on the object boundaries. 
The set of 3D points on the contours is denoted as $\contour_t := \{\contourPoint \in \mathbb{R}^3 \}$. 
Using the contours on the target image, we build the distance transform $\distanceTransform_t$.
With the predicted pose $\qpredicted$, we define the contour loss as 
\begin{align}
&\lcontours{}(\qpredicted, \distanceTransform_t, \contour_t) \nonumber \\
&:=
\sum_{\contourPoint \in \contour_t} 
\distanceTransform_t \left[ \pi \left( 
\qpredicted 
\qtrueinv
(\contourPoint - \ttrue)
\qtrue 
\qpredicted^{-1} 
+ \ttrue{} \right) \right]
\label{eq:loss_contours}
\end{align}
with $\quaternion^{-1}$ being the conjugate quaternion.
The function measures the contour's alignment after transforming the points from the target to source using the ground truth and from the source to the target using the prediction. 
Although one can argue that we can simply take the contour points on the source in order to avoid transforming back and forth, we need to consider that there are instances when the source and the target have a large relative pose between them and might have contours that do not match due to occlusions.
In this case, the contour from the source and the distance transform from the target do not match.
\begin{figure}[h]
    \begin{tabular}{ccc}
        \includegraphics[width=0.30\columnwidth, page=1]{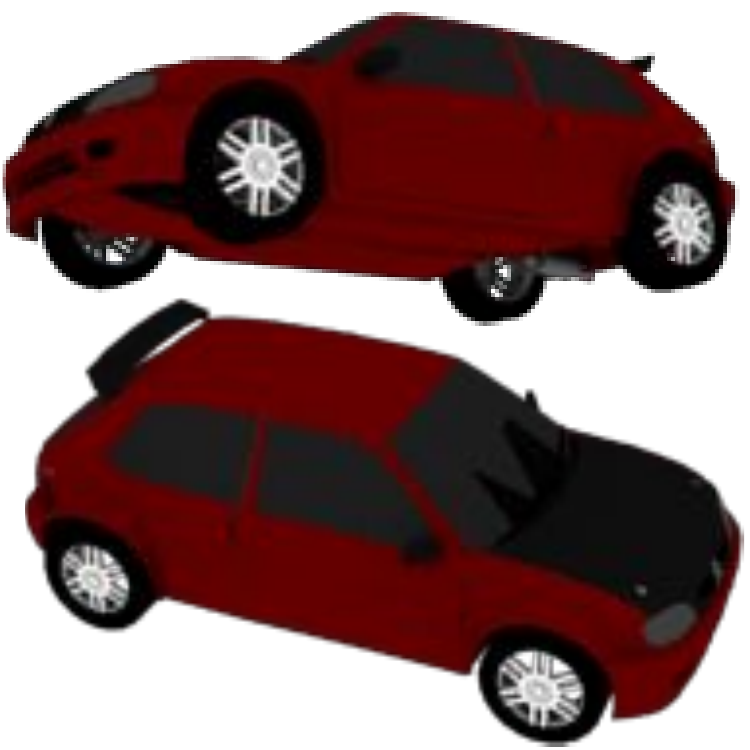} &
        \includegraphics[width=0.30\columnwidth, page=2]{images/contours_loss_failure.pdf} &
        \includegraphics[width=0.30\columnwidth, page=3]{images/contours_loss_failure.pdf} \\
        \footnotesize (a)  & \footnotesize (b)  & \footnotesize (c)  \\
        \footnotesize Input Views & \footnotesize $\lcontours{}$ &  \footnotesize $\lcontours{} + \langular{}$ \\
    \end{tabular}
    \caption{A failure case of the contour loss, on the left (a) a couple of input views and on the right (b-c) the ground truth contour (in green) and the contour oriented according to the pose estimated by the network (in red) drawn over the corresponding distance transform. The network trained only with $\lcontours{}$ (b) outputs a completely wrong pose due to ambiguity in the distance transform. Adding $\langular{}$ (c) solves the ambiguity and align the contours well.}
	\label{fig:contours_aligned}
\end{figure}

\subsubsection{Angular loss}

Despite being very effective, we found out that learning a relative pose by aligning the object contours can easily fall into local minima for objects with an elliptical structure, \textit{e.g.}~cars.
Unlike \cite{manhardt2018deep} that uses the contour loss for small pose changes between two images, \textit{i.e.}~between $4^{\circ}$ and $45^{\circ}$, the relative pose estimation in our work requires to handle large pose differences, mostly because we relax the constraint of having an overlapping area between the images. 
An example of such local minima is depicted in \autoref{fig:contours_aligned} with a rotation of almost $180^{\circ}$. 
The contour loss in this case cannot provide a useful training signal since the front and back side of the car are hardly different by just looking at the contours.

Taking advantage of the ground truth pose $\qtrue$, we can directly supervised the network with the angular difference between rotations thus establishing
\begin{equation}
\langular{}(\qtrue{}, \qpredicted{}) := 1 - \operatorname{Re}\left(\frac{\qtrue{} \qpredicted^{-1} }{ \norm{\qtrue{} \qpredicted^{-1}}}\right)
\label{eq:loss_angular}
\end{equation}
where `$\operatorname{Re}$' denotes the real part of the quaternion. 
Notably, directly regressing the pose only, with \autoref{eq:loss_angular}, confuses the network especially when dealing with symmetric objects like those in the ShapeNet dataset. 
For the same visual appearance, the ground truth rotations may be different, resulting in a one-to-many mapping as pointed out in \cite{manhardt2019explaining}. 
Using a visual proxy loss \autoref{eq:loss_contours} is a form of regularization that allows to overcome this issue.

\subsubsection{Complete loss}
Finally, the loss to train our pose estimation network
\begin{equation}
    \lpose{} = 
    \alpha \cdot \langular{} + 
    \beta  \cdot \lcontours{}
\end{equation}
is a weighted combination of \autoref{eq:loss_contours} and \autoref{eq:loss_angular}. 
In \autoref{subsec:pose_estimation_results}, we discuss the relative contribution of the two components.

\subsection{Graph-based rectification}
\label{subsec:pose_optimization}

Based on the predictions from \autoref{subsec:pose_estimation}, we can build a fully-connected graph that connects all the images through the relative poses. 
For every pair of images, we have a bi-directional link since we can predict the pose with one of them as the source while the other as target and vice versa. 
However, errors introduced from the prediction hinders us from directly using these poses for reconstruction. 
Even considering a single pair of images, there is no guarantee that the relative pose estimated for one direction is the inverse of the other. 

To solve this problem, we propose to optimize the entire graph and impose one relative pose per pair.
Taking the inspiration from the bundle adjustment~\cite{triggs1999bundle} to fix the poses using least-squares, we optimize the algebraic solution of
\begin{equation}
    \underset{\qoptimized_i}{\arg\min} 
    \sum_{i=1}^{N} 
    \sum_{\substack{j=1 \\ j \neq i}}^{N} 
    \left\|
    \rotation(\qoptimized_j) \cdot \rotation(\qoptimized_i)^{-1}  - 
    \rotation(\qpredicted_{i,j}) 
    \right\|^2
    \label{eq:pose_optimization}
\end{equation}
to find $\qoptimized_i$ which is the absolute pose of a the $i$-th view. 
Here, $\qpredicted_{i,j}$ is the predicted relative pose where the $i$-th image is the source and the $j$-th is the target.
We also convert the quaternions to rotation matrix through the function $\rotation(\cdot)$ and assume that $\qoptimized_1$ is the identity rotation that defines the reference coordinate system of the absolute poses. 
We have verified experimentally that using the norm of difference matrix in the optimization works better than the angular difference. 
A more theoretical explanation can be found in \cite{zhou2019continuity}.
Geometrically, we can interpret the difference in rotation matrices in \autoref{eq:pose_optimization} as the distance between the vectors pointing towards $x$-, $y$- and $z$-axis of the two rotations.

\subsection{Reconstruction from an occupancy grid}
\label{subsec:occupancy_grid}
From the previous stage of the pipeline, we  now have the set of images $\imageset = \{\image_i\}_{i=1}^{N}$ with the corresponding set of optimized absolute poses 
$\quaternionset = \{ \qoptimized_i\}_{i=1}^{N}$ 
with $\qoptimized_1$ as the reference.
Now, we proceed to reconstruct the object.

\begin{figure}[t]
    \centering
	\begin{tabular}{ccc}
		\includegraphics[width=0.29\linewidth, page=1]{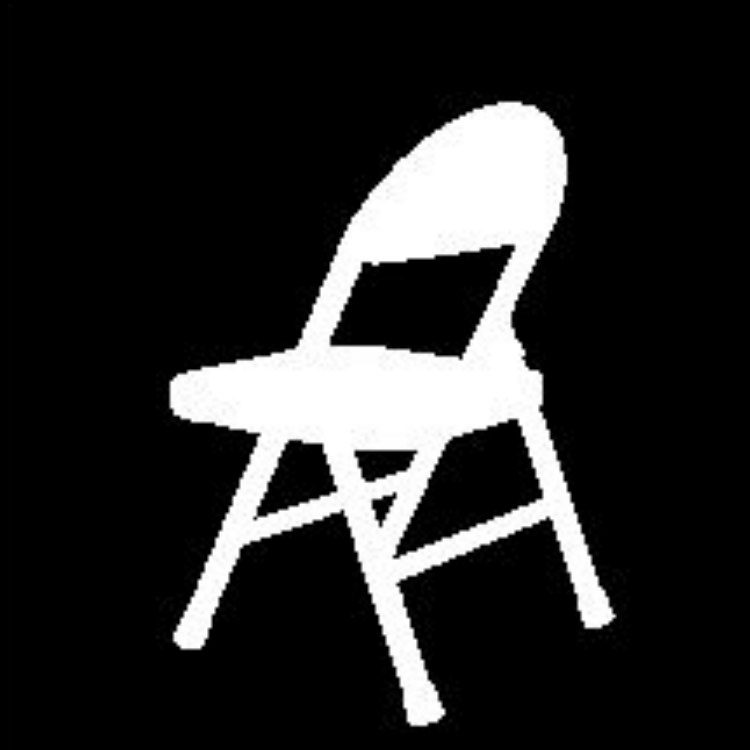} &
		\includegraphics[width=0.29\linewidth, page=2]{images/ray_casting.pdf} &
		\includegraphics[width=0.29\linewidth, page=3]{images/ray_casting.pdf}\\
		\footnotesize (a) & \footnotesize (b) & \footnotesize (c) \\
		\footnotesize Silhouette & \footnotesize Visual Cone & \footnotesize Rotated Visual Cone \\
	\end{tabular}     
	\caption{Ray casting for a single view. We show (a) silhouette of the view and (b-c) two visualization of the same visual cones -- (b) oriented according to the camera view point and (c) according to a different viewpoint.}
	\label{fig:silhouettes_ray_casting}
\end{figure}

\begin{figure*}[t]
	\centering
    \includegraphics[width=0.85\linewidth]{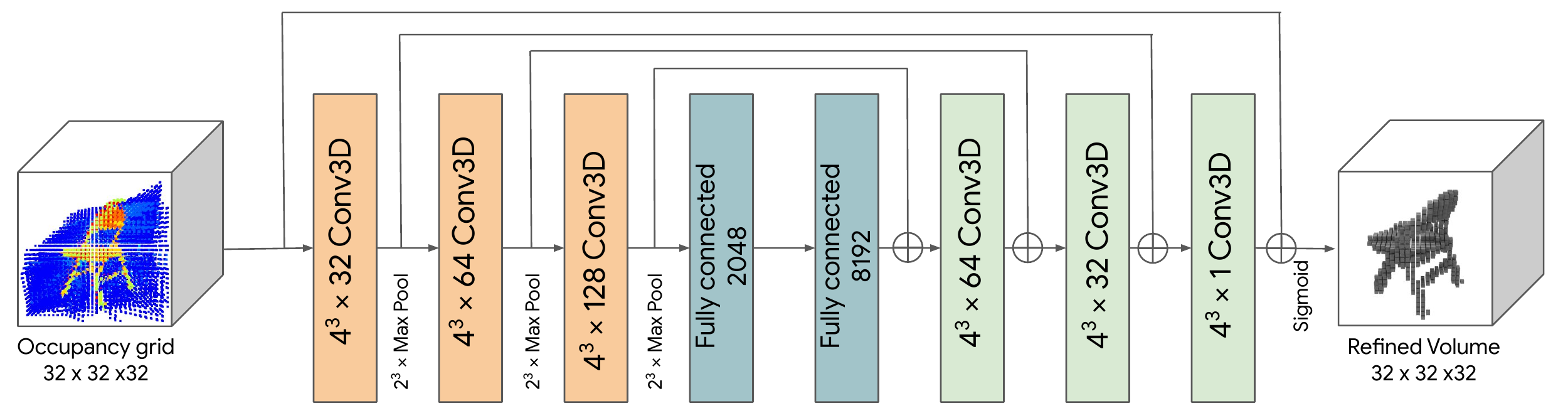}
    \caption{Architecture of the occupancy grid refiner network that predicts the final model.}
	\label{fig:refiner}
\end{figure*}

Similar to visual hull or voxel carving~\cite{laurentini1994visual}, we want to exploit the idea of shape-from-silhouette, wherein, given a 3D grid, the voxels outside the silhouette of at least one image is carved out of the grid, ending up with the shape of object.
However, directly using these methods is not feasible for two reasons. 
First, they usually require a large number of views of the object to achieve a good reconstruction. 
For instance, in \autoref{fig:silhouettes_ray_casting}, we show the problem of the reconstruction from a single view where the negative effects of ray casting become visible when the grid is rotated. 
The shape of the object becomes more detailed only when we (greatly) increase the number of views.
In our case, we must be able to also handle a handful of input images. 
Second, the errors introduced from the pose estimation may generate small misalignment among the images that may cause the carving of voxels that are on the object. 
Note that one of the most successful applications of visual hull is a multi-camera studio where all the cameras are professionally calibrated.

Instead of the classic hard thresholding used in visual hull, we propose to utilize an occupancy grid as the input to a network trained to refine this initial reconstruction. 
We denote the set of silhouette as $\silhoutteset=\{\silhouette_i\}_{i=1}^{N}$, where the pixel values are either 0 if it is outside the object or 1 if it is in the object. 
Therefore, the values of the voxels in the occupancy grid are calculated as the weighted average
\begin{equation}
    V(x) = \frac
    {\sum_{i=1}^{N} w_i \cdot S_i\left( \pi \left( 
    \qoptimized \cdot x \cdot \qoptimized^{-1} + t^*
    \right) \right)}
    {\sum_{i=1}^{N} w_i}
    \label{eq:occupancy}
\end{equation}
where $t^*$ is the ground truth camera translation, $x$ is the centroid of the specified voxel and $w_i$ is the weight assigned to the $i$-th view.  
Every voxel in the grid ranges from 0 to 1 where 0 corresponds to a voxel that is not visible from any view and 1 corresponds to a voxel that fall within the silhouette of all views.
In our implementation, we assume that the sum of all weights is one and $w_2 = w_3 = \dots = w_{N}$ while only $w_1$ changes. 
The motivation behind this assumption is to give more importance to the first view -- the reference -- since the output of the model needs to be aligned to this. 
Therefore, $w_1$ is generally weighted higher than the other weights. 
We pick as dimension of the occupancy grid $32\times 32 \times 32$.

The coarse representation of the occupancy grid can be refined with the help of deep learning a shown in \autoref{fig:refiner}.
We employed the off-the-shelf architecture proposed in Pix2Vox~\cite{Xie_2019_ICCV}  to refine the raw occupancy map and predict the reconstruction. 
Differently from Pix2Vox, we use this architecture to refine an occupancy grid while their input is a coarse voxel map fused from single-view shape reconstructions.
To generate the ground truth voxels, we reorient the 3D model with respect to the reference view and then discretize it to a grid of $32^3$ voxels.
The value of the voxels in the ground truth occupancy map is either $0$ ({free}) or $1$ ({full}), which is denoted as $\voxbintrue$. 
This is differentiated from the predicted occupancy map  $\voxbinpred$.

The network is trained to minimize the mean value of the voxel-wise binary cross entropies between the predicted occupancy map and the reoriented ground truth model 
\begin{equation}
  \lrefiner = \frac{1}{n_v} \sum_{i=1}^{n_v} 
  \left[ 
  \voxbintrue \log(\voxbinpred{}) + 
  (1 - \voxbintrue) \log(1 - \voxbinpred) \right]
  \label{eq:loss_bce}
\end{equation}
where $n_v=32^3$ is the number of voxels.
At test time, the occupancy map predicted by the network is binarized by a $0.3$ threshold to obtain the final reconstructed 3D model.
\section{Experimental results}
\label{sec:results}
We implement all the networks using Tensorflow2.0~\cite{abadi2016tensorflow} while we use Ceres~\cite{ceressolver} to optimize the poses with a Cholesky solver. 

\paragraph{Training details.}
The pose estimation and refiner models are independently trained.
To train the pose network, we render five views  for each model, compute all the possible permutations of two views and use each pair as a training example.
The network is trained with $224 \times 224$ RGB images and batch size of $24$ with the loss functions weighted by $\alpha=0.1$ and $\beta=0.9$.
We used a fixed learning rate of $0.001$ and the Adam optimizer~\cite{kingma2014adam}. 

To train the refiner network, we use the ground truth pose and silhouettes extracted from the images. 
We take a set of $5$ views and poses per model, we randomly select one view as reference then synthetically perturb the ground truth poses by a maximum of $10^\circ$ and finally build the occupancy grid.
The refiner network take  an occupancy grid of $32 \times 32 \times 32$ as input and outputs a grid with the same size. 
We train the model with the batch size of $16$, the fixed learning rate of $0.001$ and the Adam optimizer \cite{kingma2014adam}.

\paragraph{Dataset.}
The experiments are conducted using the 3D models from the ShapeNetCore.v1 dataset~\cite{chang2015shapenet}. 
Following the protocol of \cite{insafutdinov2018unsupervised,tulsiani2018multi}, we focus on the three most challenging categories: \textit{airplanes}, \textit{cars} and \textit{chairs}. 
To render the images for each object, we use the toolkit provided by \cite{tulsiani2018multi}, following the same data generation procedure.
For each model, we render five random views with random light source positions and random camera azimuth and elevation, sampled uniformly from $[0^\circ,360^\circ)$ and $[-20^\circ,40^\circ]$, respectively. 
To have a fair comparison, we use the same train and test split provided in \cite{insafutdinov2018unsupervised,tulsiani2018multi} for the pose estimation part while the split provided by \cite{choy20163d,Xie_2019_ICCV} for the shape estimation.

\paragraph{Evaluation metrics.}

To measure the pose error, we use the same metrics as Tulsiani \etal~\cite{tulsiani2018multi} which includes the accuracy, defined as the percentage of samples for which the error between the predicted pose and the ground truth is less than $30^{\circ}$, and the median error. 
For the shape estimation, we employ the Intersection Over Union (IoU) and the Chamfer Distance. 

\paragraph{Evaluation methodology.}
To evaluate the accuracy of the estimated poses, we compute the relative poses of every possible pairs of views given the ground truth absolute poses. 
Then, we directly compare the predicted relative poses with the ground truth one.

For the shape evaluation, we binarize the output of the refiner network using a threshold $\tau_r = 0.3$. 
Then, for the IoU, the ground truth voxel grid is computed using the point cloud of the 3D model oriented by the ground truth pose from the canonical view to the reference view. 
For the Chamfer distance, we convert the predicted voxel grid into a point cloud and  compute the error against the reoriented ground truth cloud uniformly sampled according to \cite{lin2018learning}.

\subsection{Pose estimation}
\label{subsec:pose_estimation_results}

We compare our relative pose estimation against MVC~\cite{tulsiani2018multi} as well as DPCD~\cite{insafutdinov2018unsupervised}. 
Both methods regress the absolute camera pose from a collection of images of the same object using a category-specific network. 
For MVC, we also report the upper bound of the method when trained with supervision (referred on the tables as \emph{GT poses}). 
We use the results reported by the authors in their respective papers. 

Notably, both competing methods need to align the canonical pose learned by the network to the canonical orientation of the dataset before starting the evaluation while we don't.
Moreover, both methods use different solutions to handle the problem of the shape similarity when looked from different camera views. 
By formulating the problem as relative pose regression, our network can easily handle these situations without using a specific mechanism.

\begin{table}
    \resizebox{\linewidth}{!}{
      \centering
       \setlength{\tabcolsep}{1.5mm}
      {\small
       \begin{tabular}{l|cccccc|cc}
       \toprule
       \multicolumn{1}{c}{Method} & \multicolumn{2}{c}{\textit{Airplane}} & \multicolumn{2}{c}{\textit{Car}}  & \multicolumn{2}{c}{\textit{Chair}} & \multicolumn{2}{c}{Mean} \\
       \midrule
       GT poses  \cite{tulsiani2018multi}             & \textbf{0.79} & 10.70 & 0.90 & 7.40 & 0.85	& 11.20 & 0.85 & 9.77 \\
       MVC \cite{tulsiani2018multi}                 & 0.69 & 14.30 & 0.87 &	5.20 & 0.81 & 7.80  & 0.79 & 9.10 \\          
       DPCD \cite{insafutdinov2018unsupervised}     & 0.75 & 8.20 & 0.86 & 5.00  & \textbf{0.86}	& 8.10  & 0.82 & 7.10 \\
       \midrule
       Ours                               & \textbf{0.79} & \textbf{6.40} & \textbf{0.93}	& \textbf{3.10} & 0.85 & \textbf{7.00} & \textbf{0.86} & \textbf{5.50}  \\
       -- \textit{without optimization}                                      & \textbf{0.79} & 6.49 & 0.92 & 3.32  & 0.85 & 7.18  & \textbf{0.86} & 5.66 \\
       -- \textit{one net all categories}                  & 0.77 & 6.80 & 0.90 & 3.40 & 0.82 & 6.80 & 0.83 & 5.70 \\
       \bottomrule
      \end{tabular}
      }
   }
  \caption{\label{tab:pose_estimation_quant} Quantitative evaluation for pose prediction. For each category, we report on the left the accuracy and on the right the median error.}
\end{table}

In \autoref{tab:pose_estimation_quant}, we report both the accuracy (on the left) and the median error (on the right) per category and averaged across all categories.
We report results for our approach with and without the optimization in \autoref{subsec:pose_optimization}.
Even considering just the raw predictions, we are already able to achieve performance better than all the competitors both in terms of the average scores and the per-category performances.
We ascribe this result to our choice of regressing relative poses between views rather than the absolute pose of a single view with respect to an implicit reference system.
The advantage of regressing the relative poses is particularly evident when comparing the median error where our method without optimization improves the state of the art by $1.44^\circ$. 
Applying the pose optimization described in \autoref{subsec:pose_optimization}, we further increase the performance obtaining the best overall results.

In the last row of \autoref{tab:pose_estimation_quant}, we report the performance of our pose estimation model when training a category agnostic network rather than a different model for each category and applying the pose optimization. 
The network needs to solve a much harder task in this case as testified by the small drop in performance ($-0.03$ in accuracy and $+0.2$ in the median error).
However, even in this case, our proposal remains competitive in terms of accuracy or lower in median error with respect to any competitor with the advantage of having a single class-agnostic model while the competitor relies on class specific models.

\paragraph{Ablation study.}
\begin{table}[tb]
	\resizebox{\columnwidth}{!}{
	\centering
	\begin{tabular}{cc|cc}
		\toprule
		~~~$\langular{}$~~~ & ~~~$\lcontours{}$~~~ & ~~~Accuracy~~~ & ~~~Median~~~ \\
		\midrule
		\xmark & \cmark & 0.45 & 78.23 \\
		\cmark & \xmark & 0.84 & 8.24 \\ 
		\cmark & \cmark & \textbf{0.86} & \textbf{5.50} \\
		\bottomrule
	\end{tabular}
	
}
	\caption{Ablation study on the angular and contour losses.}
	\label{tab:abl_pose_estimation}
\end{table}
In \autoref{tab:abl_pose_estimation}, we report an ablation study on the contribution of the two loss functions described in \autoref{subsec:pose_estimation}. 
The values of hyperparameters $\alpha$ and $\beta$ have been determined through empirical studies by considering differing combinations always summing up to 1 and measuring the performance on the Shapenet validation set.
Our model cannot achieve satisfactory performance when training with $\lcontours$ only but training with $\langular{}$ only can already provide relatively good results.
However, by mixing the two loss functions, we are able to achieve the best overall results increasing the accuracy by $0.02$ and, especially, decreasing the median error by $2.77^\circ$.

\begin{figure*}[!t]
\centering
\includegraphics[width=1.0\linewidth]{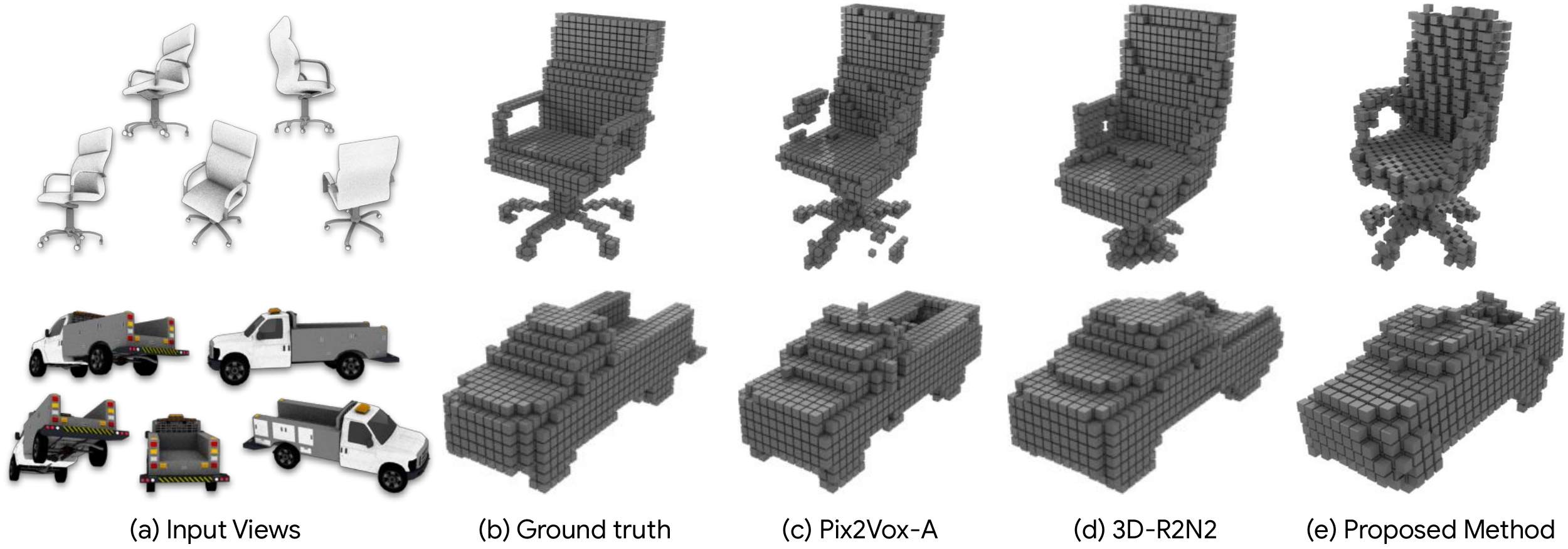}
\caption{Comparison of multi-view reconstructions methods on the ShapeNet test set for the chair and car categories. On the left, we show the five RGB views used as input for every method. We also report the results for the two main competitors Pix2Vox~\cite{Xie_2019_ICCV} and 3D-R2N2~\cite{choy20163d}.}
\label{fig:qualitative_reconstruction}
\end{figure*}

\subsection{Shape reconstruction}
\label{subsec:shape_estimation_results}
\begin{table}
    \resizebox{\columnwidth}{!}{
      \centering
       \setlength{\tabcolsep}{1.5mm}
      {\small
      \begin{tabular}{l|cccccc|cc}
      \toprule
      \multicolumn{1}{c}{Method} &\multicolumn{2}{c}{\textit{Airplane}} & \multicolumn{2}{c}{\textit{Car}}  & \multicolumn{2}{c}{\textit{Chair}} & \multicolumn{2}{c}{Mean} \\
      \midrule
      3D-R2N2 \cite{choy20163d}
      & 0.585 & 3.77 & 0.851 & 3.58 & 0.575 & 4.21 & 0.670 & 3.86 \\
      Pix2Vox \cite{Xie_2019_ICCV}
      & 0.723 & 3.20 & \textbf{0.876} & 3.54 & 0.612 & 3.77 & 0.737 & 3.50 \\
      \midrule
      Ours            
      & 0.538 & 4.81 & 0.627 & 3.93 & 0.510 & 4.75 & 0.559 & 4.50\\
      -- \textit{with GT poses}                  
      & 0.654 & 3.85 & 0.659 & 3.63 & 0.592 & 4.06 & 0.635 & 3.85 \\
      -- \textit{in Canonical} 
      & \textbf{0.732} & \textbf{2.92} & 0.874 & \textbf{3.51} & \textbf{0.648} & \textbf{3.39} & \textbf{0.751} & \textbf{3.28} \\
      \bottomrule
      \end{tabular}
      }
    }
  \caption{\label{tab:shape_estimation_merged}Quantitative evaluation for shape prediction. For each category, we report the average $\text{IoU}$ on the left and the Chamfer distance between normalized point clouds multiplied by 100.
  }
\end{table}
Given an unconstrained set of views, our method is able to reconstruct the 3D model of an object aligned with one of the views provided as input.
Unfortunately, there isn't any work in the literature that addresses the same exact settings, as structure from motion methods rely on the assumption of having overlapping views, while most 3D reconstruction methods reconstruct a model only in an arbitrary reference view.
Therefore, to provide some insightful comparison, we have chosen the most similar work in the literature as competitors -- multi-view reconstruction solutions that do not require overlapping views or pose as inputs but reconstruct the model only in a canonical reference view in the form of a voxel grid. 

In \autoref{tab:shape_estimation_merged}, we compare our work against Pix2Vox~\cite{Xie_2019_ICCV} and 3D-R2N2~\cite{choy20163d} using the publicly available implementations.
All methods take the same number of input images, \ie{} 5.
For a fair comparison, we train both the pose estimation and the refiner networks using one model for all the categories. 
From the results in \autoref{tab:shape_estimation_merged}, we can see that our approach reaches competitive results against the \sota{} proposals such as Pix2Vox \cite{Xie_2019_ICCV} and 3D-R2N2 \cite{choy20163d}. 
We argue that the slightly inferior performances are not due to a minor accurate 3D reconstruction pipeline but due to the task that we are trying to solve being harder and more realistic than the one the competitors are addressing -- \ie{} arbitrarily aligned reconstruction versus canonically aligned reconstruction.
This is true even ablating the possible errors coming from pose estimation and considering a variant of our model that takes ground truth poses as input, \ie{} \textit{Ours with GT Poses}.
For this model, the performance increases and get closer to those of the competitors but still perform slightly worse.

To highlight that the worse performance are due to the different task we are addressing and not to deficiencies of our model, we include an additional variant of our approach, \ie{} \textit{Ours in Canonical}. 
This variant uses the ground truth poses and adopts the same problem setup as the competitors by predicting an occupancy grid oriented according to the dataset's canonical pose. 
The purpose of this evaluation is twofold: verify if relying on the insight of having a canonical pose for all the train objects helps the learned model and compare only our reconstruction framework against the competitors.
The result shows how in the same testing condition our reconstruction pipeline can outperform the other methods.
Finally, in \autoref{fig:qualitative_reconstruction}, we report a visual comparison between the reconstructions obtained by our full pipeline and the competitors on the chair and cars categories of ShapeNet test set.
With the help of the estimated poses, we are able to combine the information present in the silhouettes and better keep some of the geometric structures present in the input views.
Considering the chair example, our method is the only one to correctly reconstruct the arm rests and legs of the chair, while the competing methods either produce detached parts in the models or fill parts that should be empty. 
Similarly, Pix2Vox \cite{Xie_2019_ICCV} creates a door in the trunk of the truck depicted in the second row, while 3D-R2N2 \cite{choy20163d} and our method correctly leaves it empty.

\section{Conclusions}
\label{sec:conclusion}
We have proposed a novel method for 3D reconstruction from a set of sparse and non-overlapping views.
To the best of our knowledge, our solution is the first to propose a 3D reconstruction pipeline that does not require training models from a dataset with a predefined canonical orientation and is able to reconstruct them aligned to any arbitrary input view.
This characteristic is crucial both to scale to real datasets as well as to apply this technology to common applications in the field of augmented reality and robotics.
Our pipeline is built by stand-alone components that address recurring problems in computer vision and can be easily re-used for different tasks. 
One of the main advantages of the proposed solution is to move towards a more realistic solution by relaxing the requirement of a canonical orientation for all objects, which works well for academic datasets but does not scale to real world applications.
Our method yields 3D reconstructions that are on par with the state of the art in terms of accuracy while providing a valuable additional output in terms of estimated pose from an arbitrary viewpoint. 
In the future, we plan to continue the development of our solution by relaxing the current limitations such as the fixed distance from the origin of the objects and the known instrisic camera parameters.
{\small
\bibliographystyle{ieee}
\bibliography{egbib}
}

\clearpage

\multido{\i=1+1}{3}{
	

\section*{Supplementary material}

# A Divide et Impera Approach for 3D Shape Reconstruction from Multiple Views
## (*Supplementary Materials*)

Due to the page constraints of the main submission, our supplementary material includes additional results and ablation studies to help understand the proposed method.

## 1. Shape reconstruction

In this section, we provide additional experiments on the quality of the shape reconstructed by our method. Particularly, we study how the reconstructions can be improved when more views are available. We then focus on the robustness of the refiner when the poses used to build the occupancy grid are affected by a significant amount of noise. In both experiments, we use the Chamfer Distance as metric.

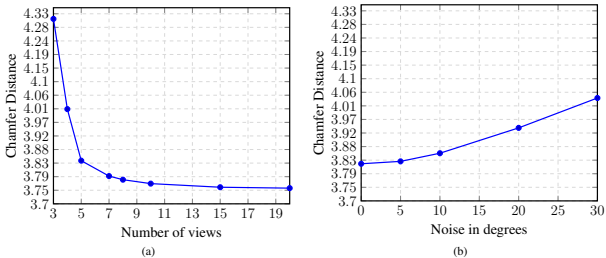

Figure 1. The Chamfer Distance computed on the ShapeNet testing set with different pre-conditions – (a) effects of using a different number of views as input; and, (b) robustness of the refiner network to the noise in the poses.

Fig. 1(a) illustrates that, as we increase the number of views when constructing the occupancy grid, the Chamfer Distance between the final reconstruction and the ground truth decreases. Increasing the number of views from 5 to 10 give us the most significant boost in the performance.

The goal of the experiment, whose results are depicted in Fig. 1(b), is to highlight how the noise in the estimated poses affects the final reconstruction. For that reason, we build the occupancy grid perturbing the ground truth poses with a random amount of noise between zero and a maximum value. We varied the value of maximum random noise between the range $0°$ to $30°$. Fig. 1(b) shows that, even with a maximum amount of noise in the poses of $15°$, the performance declines gracefully. Notably, since the median error for our pose estimation network is around $5°$, it is fair to assume that the noise in the estimated pose will not dramatically affect the final reconstruction.

## 2. Qualitative results of our pose estimation

We provide in Fig. 2 the qualitative results to show the effectiveness of our pose estimation network. We compare the results obtained by estimating the relative pose for pairs of images taken from the objects belonging to different categories in different poses.

Given the input pairs shown in the first and second columns of Fig. 2, the third column in the figure compares the point clouds of the CAD model aligned according to the ground truth pose and the predicted one, in green and red, respectively. Here, the alignment between the two point clouds verifies the accuracy we achieved in the main manuscript.

To provide a better visualization of the quality of the estimated poses, the figures in the last column encode the misalignment error computed as the per point Euclidean Distance between the point cloud rotated according to the ground truth matrix and the point cloud oriented with the predicted one. Note that the errors are normalized according to the maximum error. From these results, we can observe that our network produces good alignments even in cases where the rotation between the source and target image is large such as the first, fourth and fifth row (low overlapping area between views). This is clearly evident in the point clouds in column (c) that almost completely overlap and the misalignment error in column (d) which is consistently low for most points.

## 3. Qualitative results of our shape reconstruction

In Fig. 3, we show more qualitative results of the reconstruction obtained by our pipeline for the three categories considered: airplanes, cars and chairs.

Due to the *view-dependent* reconstruction, the voxel grid obtained to supervise our model in training has a lower spatial resolution compared to the voxel grid available in the ShapeNet dataset [1]. Indeed, to get a voxel grid w.r.t.

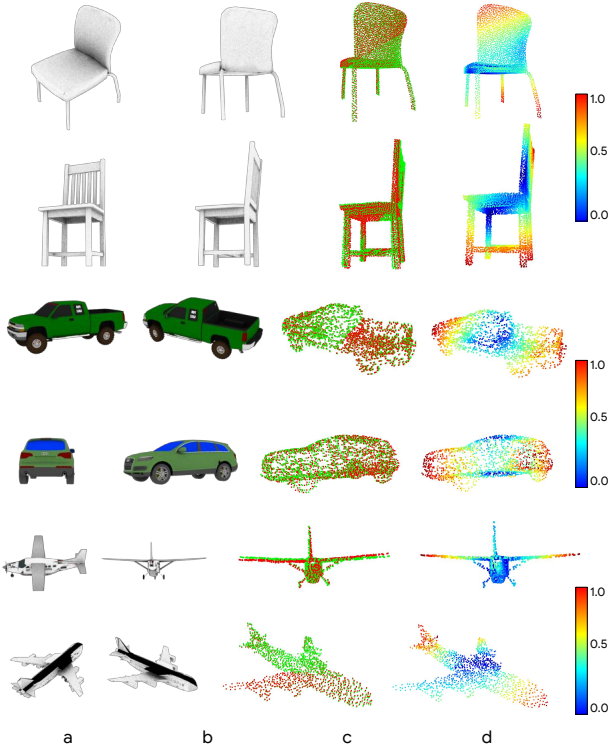

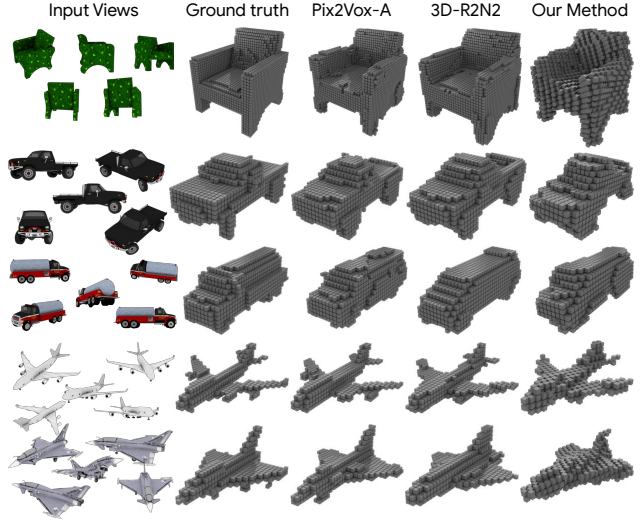

Figure 3. Comparison of multi-view reconstructions methods on the ShapeNet test set for the chair, car and plane category. On the left, we show the 5 RGB views used as input for every method. We also report the results for the two main competitors Pix2Vox [3] and 3D-R2N2 [2].

Figure 2. Qualitative results for the pose estimation network. We show on the left (a, b) the source and target input image, on the right (c) the CAD model point cloud, in *green*, oriented according to the ground truth pose, and the same point cloud, in *red*, oriented with the pose predicted by our method. In the last column (d), we visualize the normalized misalignment error between the two models as a heat map ranging from blue (perfect alignment) to red (maximum misalignment).

the reference view, we first align the point cloud of the CAD model to the reference view, then we voxelize it in a $32 \times 32 \times 32$ grid. As a result, the predicted shapes have a smaller spatial resolution since they need to be aligned with the input RGB images, and therefore our refiner model outputs a shape with the same lower spatial resolution. This difference is even more clear if we look at the reconstruction in Sec. 4 where *Ours Canonical* is trained using instead the voxel grid available in ShapeNet.

Considering the reconstruction of chairs, trucks and planes in Fig. 3, our method correctly approximate the shape of the models while, at the same time, keeps some of the fine details like the flash light and the rear floor for the truck in the second row.

## 4. Qualitative Results using the canonical orientation

This section compares the generated shapes of two variants of our reconstruction pipeline – the standard one reconstructing in an arbitrary reference frame aligned with one of the input views (*Ours*); and, the one that always reconstruct a model in a canonical reference frame (*Ours Canonical*). The purpose of this comparison is to clarify the performance of the second stage of our method, which is composed of building the occupancy grid and the refiner network. For *Ours Canonical*, we first build an occupancy grid oriented as the canonical orientation of the ShapeNet dataset, then we refine this volume using the refiner network. As already pointed out in section Sec. 3, to supervise our network, for (*Ours Canonical*), we use as ground truth the voxel grid available from the ShapeNet dataset. When comparing the reconstruction of *Ours Canonical* to *Ours*, we can see how the reconstruction on a fixed reference frame results in more detailed and smooth models. We ascribe this difference to the task learned being simpler than reconstruction w.r.t. an arbitrary viewpoint. When comparing the reconstructions of *Ours Canonical* to those of the competing methods, we can see once again how our method is able to obtain a similar quality of the reconstructed shape, while, at the same time, maintaining the fine details like: the legs and arm rests in the chair, the exhaust pipes in the truck or the propellers in the plane. We believe that these results clearly show how our 3D reconstruction from silhouettes projected in an occupancy grid is as effective (or more) than the alternatives proposed in the literature. However, reconstruction w.r.t. an arbitrary reference frame is a much harder task and this is reflected in slightly less detailed models for *Our* pipeline without the canonical frame reconstruction. We will provide more details on how the reconstruction change with respect to the view choose as reference in Sec. 5

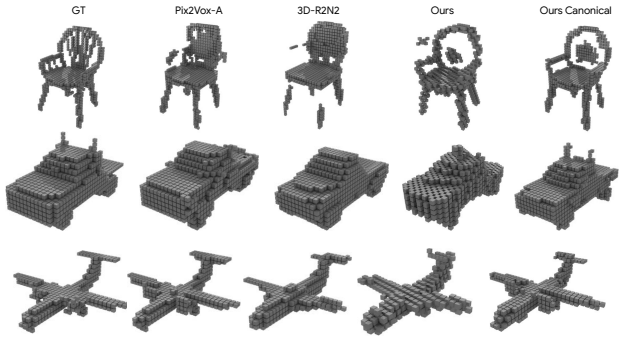

Figure 4. Comparison of multi-view reconstructions methods on the ShapeNet test set. On the left, we show the ground truth voxel grid. We also report the results for the two main competitors Pix2Vox [3] and 3D-R2N2 [2].

## 5. Qualitative results when changing the reference view

Here, we show how our pipeline is able to reconstruct different models with respect to the reference view considered when building the occupancy grid. First of all, we would like to point out that our method can correctly reconstruct a model that is nicely aligned to each of the views when it is selected as reference. Secondly, we want highlight how the fine details on the reconstructions depends on how well the corresponding details are visible in the original reference image. For example, the reconstruction of the legs of the first chair is more detailed when selecting as reference view one of the two rightmost one where the geometrical structure of the leg is cleanly visible. The same consideration can be extended to the second chair where the shape that mostly highlights the peculiar shape of the backrest is the one obtained when considering as reference the third view.

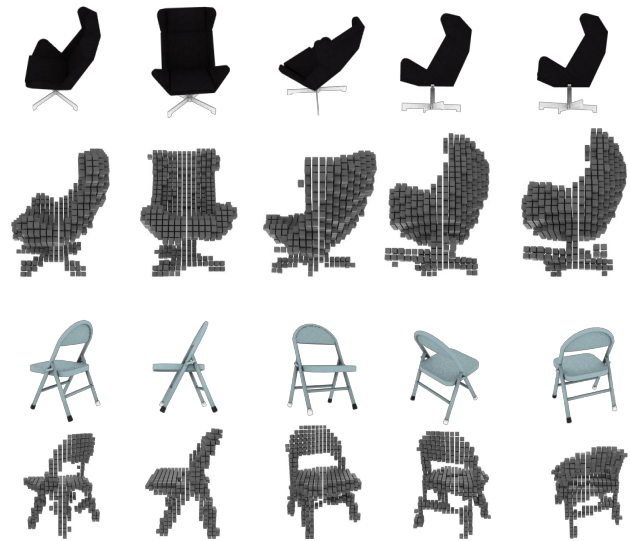

Figure 5. Comparison of multi-view reconstructions methods on the ShapeNet test set for the chair category. In each column, we depict the voxel grid produced using a different reference view.


}

\end{document}